\pdfoutput=1

\documentclass[11pt]{article}

\usepackage[]{acl}

\usepackage{times}
\usepackage{latexsym}

\usepackage[T1]{fontenc}

\usepackage[utf8]{inputenc}

\usepackage{microtype}

\usepackage{inconsolata}

\usepackage{graphicx}
\usepackage{subcaption}
\usepackage{booktabs}
\usepackage{multirow}

\usepackage{amsmath,amsfonts,bm}

\def\eqref#1{equation~\ref{#1}}

\def\1{\bm{1}}

\def\vt{{\bm{t}}}

\DeclareMathAlphabet{\mathsfit}{\encodingdefault}{\sfdefault}{m}{sl}
\SetMathAlphabet{\mathsfit}{bold}{\encodingdefault}{\sfdefault}{bx}{n}

\def\sA{{\mathbb{A}}}

\def\sP{{\mathbb{P}}}
\def\sQ{{\mathbb{Q}}}

\def\sT{{\mathbb{T}}}

\def\sX{{\mathbb{X}}}

\def\sZ{{\mathbb{Z}}}

\title{Thread of Thought Unraveling Chaotic Contexts
}

\author{Yucheng Zhou$^{1}$\thanks{~~Work is done during internship at Microsoft.} ,
        Xiubo Geng$^{2}$, Tao Shen$^{3}$, Chongyang Tao$^{2}$, \\ {\bf Guodong Long}$^{3}$, {\bf Jian-Guang Lou}$^{2\dag}$, {\bf Jianbing Shen}$^{1}$\thanks{~~Corresponding author.}\\
         $^{1}$ SKL-IOTSC, CIS, University of Macau, \\
         $^{2}$Microsoft Corporation, $^{3}$AAII, FEIT, University of Technology Sydney \\
         {\tt yucheng.zhou@connect.um.edu.mo, \{xigeng,chongyang.tao,jlou\}@microsoft.com } \\
         {\tt \{tao.shen, guodong.long\}@uts.edu.au, jianbingshen@um.edu.mo}
         }

\begin{document}
\maketitle
\begin{abstract}
Large Language Models (LLMs) have ushered in a transformative era in the field of natural language processing, excelling in tasks related to text comprehension and generation. Nevertheless, they encounter difficulties when confronted with chaotic contexts (e.g., distractors rather than long irrelevant context), leading to the inadvertent omission of certain details within the chaotic context. In response to these challenges, we introduce the ``Thread of Thought'' (ThoT) strategy, which draws inspiration from human cognitive processes. ThoT systematically segments and analyzes extended contexts while adeptly selecting pertinent information. This strategy serves as a versatile ``plug-and-play'' module, seamlessly integrating with various LLMs and prompting techniques. In the experiments, we utilize the PopQA and EntityQ datasets, as well as a Multi-Turn Conversation Response dataset (MTCR) we collected, to illustrate that ThoT significantly improves reasoning performance compared to other prompting techniques.
\end{abstract}

\section{Introduction}
Large Language Models (LLMs) represent a significant advancement in the field of artificial intelligence. They have achieved notable accomplishments in natural language understanding and generation \cite{BrownMRSKDNSSAA20,Wei0SBIXCLZ22CoT}. The development of LLMs has had a far-reaching impact, drawing significant attention in academia. These models demonstrate proficiency in a wide array of natural language processing tasks, including sentiment analysis \cite{Zhang2305Sentiment}, machine translation \cite{MoslemHKW23}, and summarization \cite{TamMZKBR23}. Moreover, they exert a profound influence across various industries and offer promising solutions for intricate issues, such as aiding in legal consultations \cite{Yue2309LawLLM} and assisting in medical diagnostics \cite{Wang2302ChatCAD}.

\begin{figure*}
    \centering
    \includegraphics[width=1\linewidth]{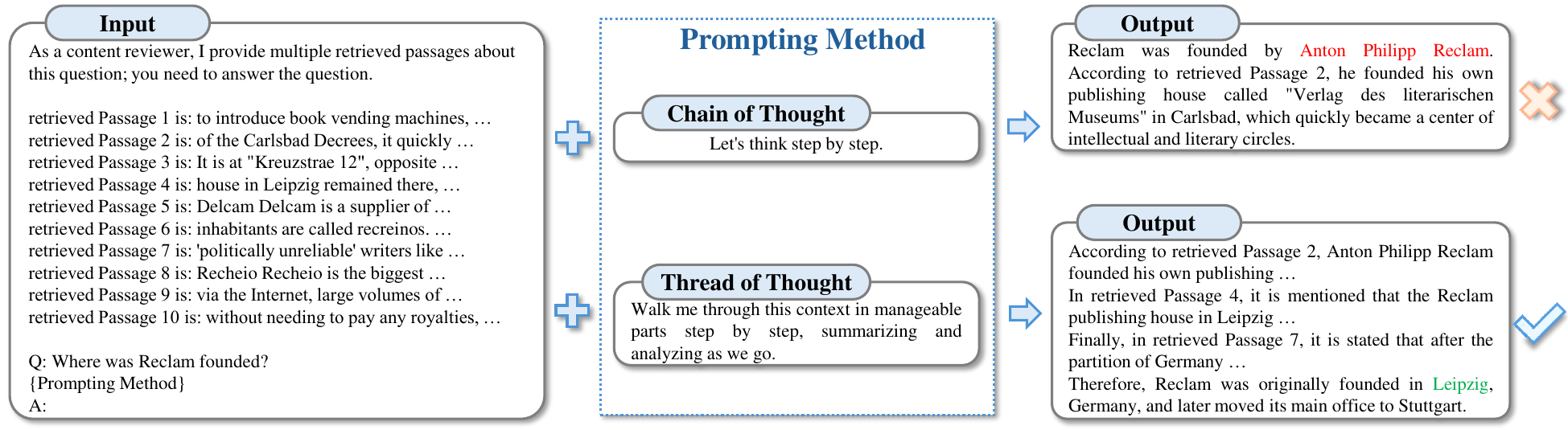}
    \caption{Thread of Thought prompting enables large language models to tackle chaotic context problems. In the output depicted, green text denotes the correct answer, while red text indicates the erroneous prediction.}
    \label{fig:intro}
\end{figure*}

With the growing complexity and diversity of tasks demanding extensive information processing and reasoning, particularly in the context of Retrieval-Augmented Generation (RAG) \cite{LewisPPPKGKLYR020} and conversational \cite{XuSW22} scenarios, the input text often comprises a wealth of information from various sources, including user queries, conversation history, external knowledge bases, and more. This information may be interconnected or entirely unrelated. Moreover, the significance of this information can fluctuate based on the context, with certain pieces being critical for addressing specific questions and others being extraneous. This situation can aptly be characterized as a ``Chaotic Context''. Similar to but distinct from ``Long Context'', ``Chaotic Context'' underscores the complexity and volume of information, going beyond the mere length of the context. Moreover, \citet{Liu2307LostinMiddle} found that existing LLMs often encounter difficulties in effectively identifying relevant information from the context augmented through retrieval, particularly when it is located in the middle position.

Recent studies \cite{Xu2310Retrieval,jiang2023longllmlingua} have proposed various solutions to enhance the performance of LLMs in long-context scenarios, upon an intuition of input capacity optimization. 
\citet{Xu2310Retrieval} proposes a method to compare and combine retrieval-augmentation and long context extension for large language models (LLMs) on various long text tasks. However, this approach necessitates the retraining of LLMs with positional interpolation.
In addition, \citet{jiang2023longllmlingua} introduces LongLLMLingua, a method that streamlines the input prompt by culling irrelevant or redundant information. Nonetheless, this method mandates the fine-tuning of auxiliary models (e.g., LLaMA-7B \cite{Touvron2302LLaMA}) for prompt compression. The utility of these auxiliary models may prove insufficient in addressing unknown or intricate content, and it imposes limitations on the length of text that can be effectively processed. Moreover, its non-end-to-end framework can lead to error propagation. In contrast, chain-of-thought (CoT) prompting \cite{Wei0SBIXCLZ22CoT} can enhance a model's reasoning ability without requiring any retraining or fine-tuning of LLMs. 
However, due to the massive amount of information contained within chaotic contexts, CoT still encounters information missing in reasoning, as shown in Figure \ref{fig:intro}.

To address these challenges, we introduce the ``Thread of Thought'' (ThoT) strategy. ThoT, drawing inspiration from human cognitive processes, enables Large Language Models (LLMs) to methodically segment and analyze extended contexts. This segmentation enhances the extraction of pertinent content for responding to queries. ThoT represents the unbroken continuity of ideas that individuals maintain while sifting through vast information, allowing for the selective extraction of relevant details and the dismissal of extraneous ones. This balance of attention across a document's sections is crucial for accurately interpreting and responding to the information presented. Moreover, the stepwise analysis and summarization of segmented information improve comprehension over multiple paragraphs and protect LLMs against misleading yet seemingly relevant data.

In comparison to existing methods that require complex multi-stage prompting \cite{ZhouSHWS0SCBLC23} or multi-path sampling \cite{0002WSLCNCZ23}, ThoT is a simpler, more universal, and efficient solution. It integrates seamlessly as a ``plug-and-play'' module with various pre-trained language models and prompting strategies, avoiding complex procedures. ThoT not only improves LLMs' performance in chaotic contexts but also enhances their reasoning abilities.

To evaluate ThoT's effectiveness in handling chaotic contextual information, we used long-tail question answering datasets, specifically PopQA \cite{MallenAZDKH23} and EntityQ \cite{SciavolinoZLC21}. These datasets feature knowledge often unfamiliar to large models, thereby reducing the impact of their inherent knowledge retention on our results. Additionally, we construct a Multi-Turn Conversation Response (MTCR) dataset based on everyday conversations to further assess our method. Comparative analyses with other prompting techniques show that ThoT markedly improves reasoning performance, evidencing its effectiveness. We also explored various prompts to determine optimal prompting strategies.

\section{Related Work}
\subsection{Long Context Large Language Models}
Recent advancements in Large Language Models (LLMs) have made significant strides in managing extended contexts, moving beyond the limitations of traditional pre-defined context windows. \citet{RatnerLBRMAKSLS23} introduce the Parallel Context Windows (PCW) method, which segments extensive contexts into multiple windows, employing independent attention mechanisms. Building on this concept, \citet{Chen2306Extending} facilitate substantially longer context windows with minimal fine-tuning by aligning position indices with the maximum position index from the pre-training phase. Moreover, a different approach, LongNet, utilizes dilated attention, allowing the attention field to expand exponentially with distance \cite{Ding2307LongNet}. In addition, \citet{Xiao2309Efficient} underscore the phenomenon of attention convergence, where maintaining the Key-Value (KV) states of initial tokens significantly enhances window attention performance. Lastly, \citet{PressSL22} introduce Attention with Linear Biases (ALiBi), a method that biases the query-key attention scores based on distance, achieving comparable perplexity to models trained on longer sequences. However, these methods predominantly concentrate on long contexts. In contrast, chaotic contexts are characterized by their overloaded information, often cluttered with numerous similar and unrelated elements.

\subsection{Reasoning with Large Language Models}
Advancements in large language models (LLMs) have significantly impacted AI, notably in complex reasoning tasks. The enhancement of LLMs' reasoning capabilities is exemplified in \cite{Wei0SBIXCLZ22CoT}, where Chain-of-Thought (CoT) prompting is introduced. This method improves arithmetic, common sense, and symbolic reasoning by generating intermediate steps. Building on this, the Graph of Thoughts (GoT) framework conceptualizes LLM outputs as graphs, leading to notable improvements in task performance and efficiency \cite{Besta2308got}. Extending the CoT concept, \citet{Yao2305tot} propose the Tree of Thoughts (ToT) framework, which has shown remarkable success in complex problem-solving tasks like the 24-point game. In addition, \citet{ZhouSHWS0SCBLC23} introduce the least-to-most prompting strategy, breaking down complex problems into simpler sub-problems and showing effectiveness in tasks requiring advanced symbolic manipulation. Lastly, \citet{Yao2305Beyond} explore non-linear thought processes through GoT reasoning, outperforming the linear CoT approach in both mathematical and financial problem datasets. However, these methods are effective but overlook chaotic context scenarios.

\subsection{Knowledge Following in Long Context}
LLMs can process extensive input contexts, but their performance significantly deteriorates when extracting relevant information buried in these contexts, challenging their efficiency in managing long contexts \cite{Liu2307LostinMiddle}. To address deploying LLMs in streaming applications, \citet{Xiao2309Efficient} introduce the StreamingLLM framework, enabling LLMs with limited attention windows to handle indefinitely long sequences without additional fine-tuning. Some study finds that retrieval augmentation enables a 4K context window LLM to equal the performance of a 16K context window LLM fine-tuned with positional interpolation in long-context tasks, underscoring the potential of retrieval methods in augmenting LLM capabilities \cite{Xu2310Retrieval}. Moreover, LongLLMLingua introduces prompt compression to improve LLMs' key information perception, significantly boosting performance \cite{jiang2023longllmlingua}.

\section{Methodology}
\begin{figure}
    \centering
    \includegraphics[width=0.85\linewidth]{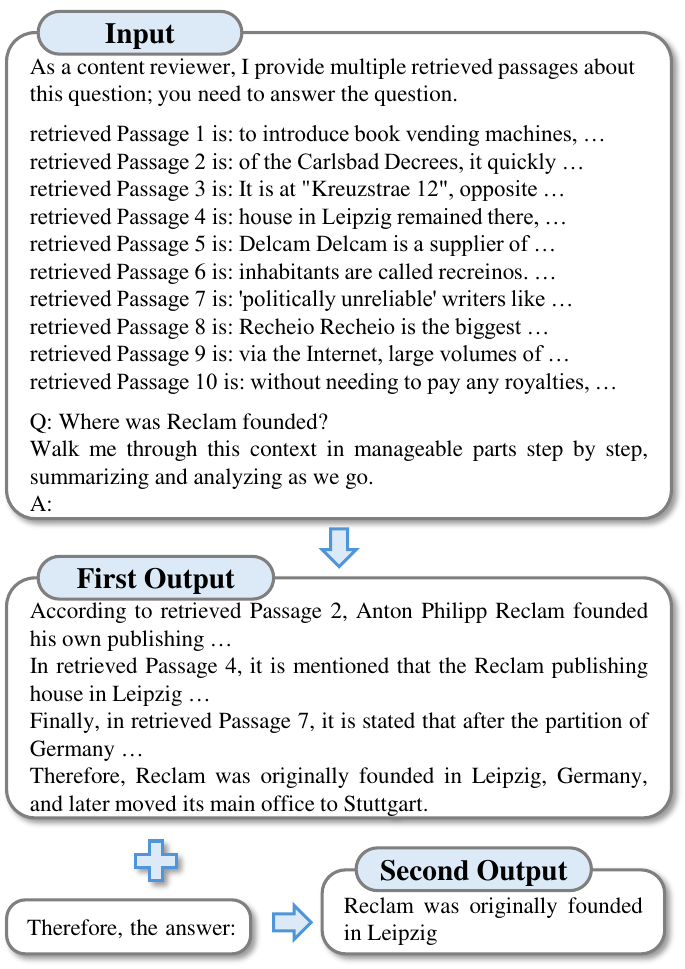}
    \caption{Thread of Thought for zero-shot reasoning.}
    \label{fig:method}
\end{figure}
We present an innovative method for template-based prompting that is specifically designed to enhance Thread of Thought (ThoT) reasoning. This novel strategy stands distinct from the traditional chain of thought prompting \cite{Wei0SBIXCLZ22CoT}, adept at navigating through disordered contexts in which the information may be either interwoven or disparate. ThoT prompting can be seamlessly integrated with a variety of existing language models and prompting techniques, offering a modular ``plug-and-play'' improvement that eliminates the need for elaborate prompting strategies or sampling methods. Our approach's underlying principle is both simple and efficient, as exemplified in Figure \ref{fig:method}: inserting ``Walk me through this context in manageable parts step by step, summarizing and analyzing as we go'' into the prompt facilitates ThoT reasoning.

As illustrated in Figure \ref{fig:method}, in contrast to Chain of Thought (CoT) prompting, which struggles with complex and chaotic contexts, ThoT prompting adeptly maintains the logical progression of reasoning without being overwhelmed. While prompt compressors and similar strategies have sought to address these complexities, they often underperform with unfamiliar or particularly complex material and typically necessitate significant modifications to the Large Language Models (LLMs), such as retraining or fine-tuning with additional datasets \cite{Xu2310Retrieval,jiang2023longllmlingua}. ThoT, however, not only effectively manages chaotic contexts but also simplifies the prompting process, requiring just two prompting efforts compared to CoT.

\subsection{First Step: Initiating the Reasoning}
The initial prompt is designed to guide the LLM through an analytical dissection of the context, using the directive ``Walk me through this context in manageable parts step by step, summarizing and analyzing as we go''. Specifically, we employ a template that incorporates the chaotic context $\sX$ and query $\sQ$ into the prompt $\sP$ as ``[$\sX$] Q: [$\sQ$] [$\sT$] A:'', where $[\sT]$ denotes the trigger sentence $\vt$ that initiates the reasoning process. For instance, utilizing ``Walk me through this context in manageable parts step by step, summarizing and analyzing as we go'' as the trigger, the prompt $\sP$ becomes ``[$\sX$] Q: [$\sQ$] Walk me through this context in manageable parts step by step, summarizing and analyzing as we go. A:''. This prompted text $\sP$ is then inputted into an LLM, which generates the subsequent sentences $\sZ$. This procedure is modeled after the cognitive strategies humans employ when confronted with complex information, breaking it down into digestible segments, distilling key points, and navigating through the material with sustained focus. This incremental method fosters a more structured and coherent line of reasoning, proving particularly advantageous in chaotic contexts.

\subsection{Second Step: Refining the Conclusion}
The second prompt builds upon the structured reasoning established earlier, employing another prompt to distill the analysis into a definitive answer. By leveraging the organized thought sequence initiated by the first prompt, this step aims to succinctly capture the essence of the conclusion. Specifically, we use a simple template to combine the initial prompted text $\sP$, the response $\sZ$, and the conclusion marker $[\sA]$, as in ``[$\sP$] [$\sZ$] $[\sA]$'', where $[\sA]$ signifies the trigger sentence designed to extract the answer, such as ``Therefore, the answer:''. This extraction prompt perpetuates the thought process, prompting the model to sift through the analysis and isolate the principal conclusion as the final answer. The prompt's design is a deliberate tactic to sharpen the model's focus, fostering precision and explicitness in the response.

This two-tiered prompting system effectively addresses the limitations of prior methods while obviating the need for intensive model retraining or complex modifications. Our methodology not only enhances the model's capacity to navigate chaotic contexts but also more closely aligns its reasoning processes with human cognitive patterns.

\section{Experiments}
\subsection{Experimental Settings}
\begin{figure}
    \centering
    \includegraphics[width=1\linewidth]{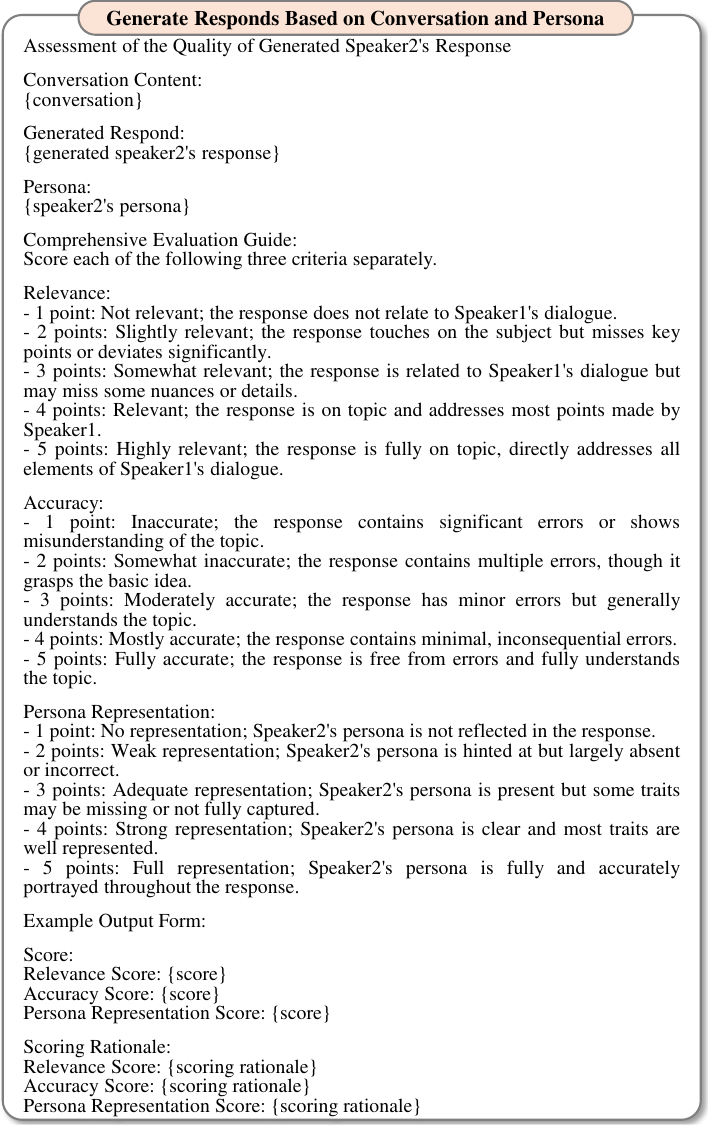}
    \caption{Prompt Evaluation Metric for MTCR Dataset.}
    \label{fig:metric}
\end{figure}

\paragraph{Dataset.}
\begin{figure}
    \centering
    \includegraphics[width=0.9\linewidth]{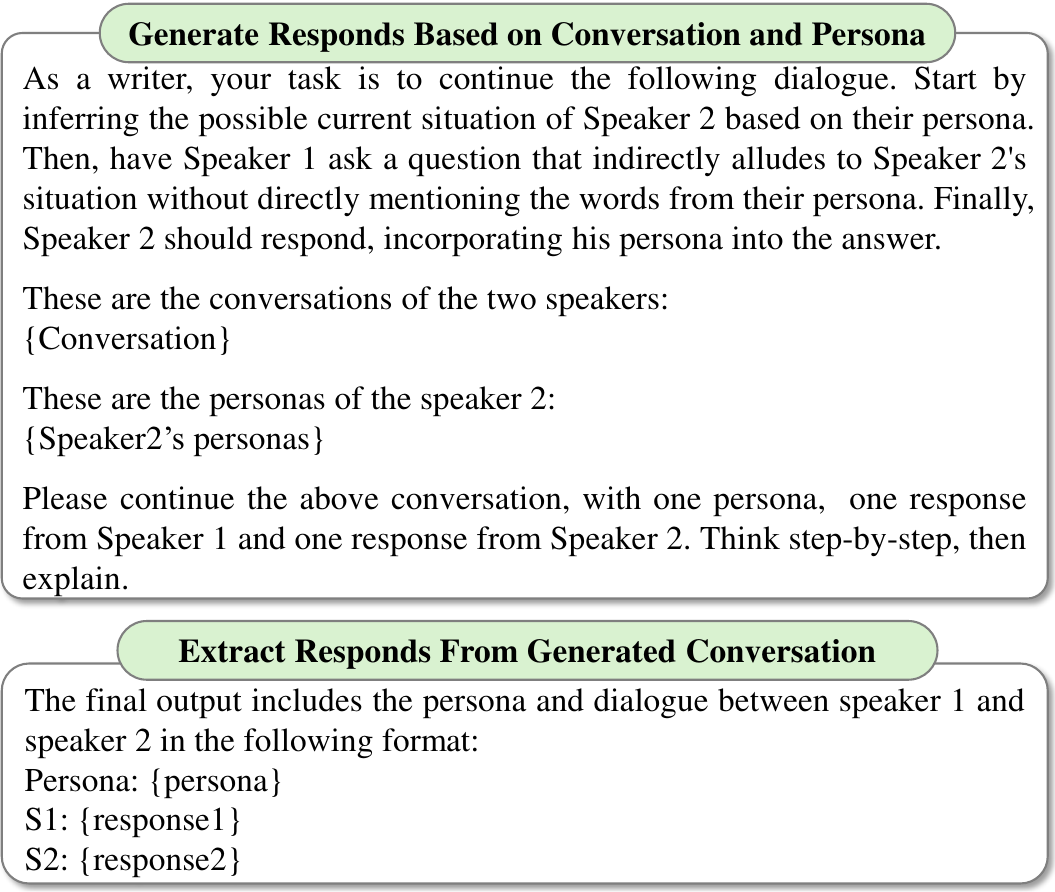}
    \caption{Prompt for MTCR Dataset Construction.}
    \label{fig:mtcr}
\end{figure}
We evaluated our method across two chaotic context scenarios: retrieval-augmented generation and multi-turn conversation response. Our assessment utilized three datasets: the PopQA dataset \cite{MallenAZDKH23}, the EntityQ dataset \cite{SciavolinoZLC21}, and our own Multi-Turn Conversation Response (MTCR) dataset. Specifically, the PopQA and EntityQ datasets, designed to contain long-tail knowledge, were chosen to minimize interference from the extensive internal knowledge of large models, thereby facilitating a more effective comparison of different methodologies. Distinct from the original PopQA and EntityQ datasets, we randomly selected a test set of 1,000 samples for our analysis. For the evaluation of the PopQA and EntityQ datasets, we adhered to the original datasets' metric, namely the exact match (EM).	
Furthermore, the MTCR dataset, used to assess multi-turn conversation response, was developed based on the Multi-Session Chat (MSC) dataset \cite{XuSW22}. The dataset construction involved sequentially using two prompts, as shown in Figure \ref{fig:mtcr}. The input of prompts is the MSC dataset's conversation and Speaker2's persona to generate a response for Speaker1. During the inference phase, the model was required to consider the multi-turn conversation contextual details mentioned previously to generate a response for speaker2, coping with the response created for speaker1. Following this, a manual screening process was conducted to eliminate samples that did not meet certain criteria, such as persona content leakage and irrelevance to the context or persona, culminating in a refined selection of 304 samples. For the MTCR dataset's evaluation, we merge the persona as a known condition along with the model-generated response for Speaker2 in the prompt, as depicted in Figure \ref{fig:metric}, and then pass them into GPT-4 \cite{GPT4}, obtaining scoring.

\paragraph{Prompt.}
In the experimental comparison, we consider four distinct prompts for retrieval-augmented generation. (1) ``Vanilla'' entails using the instruction and question as the prompt without providing any retrieval results, i.e., ``\{instruction\} \{question\}.''. (2) ``Retrieval'' includes retrieval results within the prompt, formatted as ``\{instruction\} \{retrieval results\} \{question\}.''. (3) ``CoT'' (Chain of Thought) incorporates the retrieval results and appends the phrase ``Let's think step by step'' to the instruction and question, resulting in ``\{instruction\} \{retrieval results\} \{question\} Let's think step by step.''. (4)``ThoT'' (Thought-by-Thought) also integrates retrieval results and follows a more detailed prompt structure: ``\{instruction\} \{retrieval results\} \{question\} Walk me through this context in manageable parts step by step, summarizing and analyzing as we go.''. For the MTCR dataset, we employ only the ``Vanilla'', ``CoT'', and ``ThoT'' prompts. Their formats are, respectively: ``\{instruction\} \{conversation\}'', ``\{instruction\} Let's think step by step. \{conversation\}'', and ``\{instruction\} Walk me through this context in manageable parts step by step, summarizing and analyzing as we go. \{conversation\}''.

\paragraph{Language models.}
We evaluated four large-scale language models: GPT-3.5-turbo \cite{schulman2022chatgpt}, GPT-4 \cite{GPT4}, LLaMA 2 Chat \cite{llama2}, and Vicuna \cite{vicuna2023}. Due to the GPT-3.5-turbo and GPT-4 are not open-source, the details of their model parameters remain undisclosed. For the LLaMA 2 Chat model, we utilized variants with 7B, 13B, and 70B parameters in our experiments. Similarly, versions with 7B, 13B, and 33B parameters of the Vicuna model were employed. Sampling from these models was conducted using a greedy decoding strategy.
\begin{table}[!t]\small
\centering
\begin{tabular}{lcc}
\toprule
\bf Method    & \bf GPT-3.5-turbo  & \bf LLaMA 2 Chat (70B) \\
\midrule
Vanilla   & 0.398          & 0.330              \\
Retrieval & 0.475          & 0.510              \\
CoT       & 0.482          & 0.525              \\
ThoT      & \textbf{0.574} & \textbf{0.561}     \\
\bottomrule
\end{tabular}
\caption{Performance Comparison on PopQA.}
\label{tab:popqa}
\end{table}

\begin{table}[!t]\small
\centering
\begin{tabular}{lcc}
\toprule
\textbf{Method} & \textbf{GPT-3.5-turbo} & \textbf{LLaMA 2 Chat (70B)} \\
\midrule
Vanilla         & 0.497                  & 0.430                       \\
Retrieval       & 0.512                  & 0.522                       \\
CoT             & 0.517                  & 0.547                       \\
ThoT            & \textbf{0.565}         & \textbf{0.559}              \\
\bottomrule
\end{tabular}
\caption{Performance Comparison on EntityQ.}
\label{tab:entityq}
\end{table}

\begin{table*}[!t]\small
\centering
\begin{tabular}{lcccccccc}
\toprule
\multirow{2}{*}{\textbf{Method}} & \multicolumn{4}{c}{\textbf{GPT-3.5-turbo}}                 & \multicolumn{4}{c}{\textbf{LLaMA 2 Chat (70B)}}            \\ \cmidrule(lr){2-5} \cmidrule(lr){6-9}
                                 & \textbf{Relevance} & \textbf{Accuracy} & \textbf{Persona} & \textbf{Average} & \textbf{Relevance} & \textbf{Accuracy} & \textbf{Persona} & \textbf{Average} \\\midrule
Vanilla                          & 3.211              & 3.135             & 3.345            & 3.230            & 2.819              & 2.901             & 2.914            & 2.878            \\
CoT                              & 3.352              & 3.220             & 3.349            & 3.307            & 2.783              & 2.806             & 2.882            & 2.823            \\
ThoT                             & \textbf{3.849}     & \textbf{3.921}    & \textbf{3.645}   & \textbf{3.805}   & \textbf{3.158}     & \textbf{3.295}    & \textbf{3.268}   & \textbf{3.240} \\
\bottomrule
\end{tabular}
\caption{Performance Comparison on MTCR dataset.}
\label{tab:mtcr}
\end{table*}

\begin{table*}[!t]\small
\centering
\begin{tabular}{lcccccc}
\toprule
\multicolumn{1}{c}{\multirow{2}{*}{\textbf{Method}}} & \multicolumn{3}{c}{\textbf{PopQA}}                                    & \multicolumn{3}{c}{\textbf{EntityQ}}                                  \\ \cmidrule(lr){2-4} \cmidrule(lr){5-7}
\multicolumn{1}{c}{}                                 & \textbf{GPT-4} & \textbf{GPT-3.5-turbo} & \textbf{LLaMA 2 Chat (70B)} & \textbf{GPT-4} & \textbf{GPT-3.5-turbo} & \textbf{LLaMA 2 Chat (70B)} \\
\midrule
Vanilla                                              & 0.430          & 0.391                  & 0.314                       & 0.405          & 0.405                  & 0.369                       \\
Retrieval                                            & 0.360          & 0.477                  & 0.430                       & 0.571          & 0.560                  & 0.643                       \\
CoT                                                  & 0.442          & 0.465                  & 0.558                       & 0.560          & 0.583                  & 0.667                       \\
ThoT                                                 & \textbf{0.651} & \textbf{0.674}         & \textbf{0.663}              & \textbf{0.643} & \textbf{0.667}         & \textbf{0.702}              \\
\bottomrule
\end{tabular}
\caption{Study of ``Lost in Middle'' in PopQA and EntityQ.}
\label{tab:middle}
\end{table*}
\subsection{Results}
Tables \ref{tab:popqa} and Tables \ref{tab:entityq} show the performance of retrieval-augmented generation. In PopQA and EntityQ datasets, we notice a consistent pattern where the Thought-by-Thought (ThoT) prompt configuration outperforms the other methods. The introduction of CoT also demonstrates a positive effect, indicating that prompting models to follow a methodical problem-solving approach can improve performance metrics. It is particularly noteworthy that ThoT exhibits a marked improvement in results over the CoT configuration, highlighting the efficacy of stepwise contextual processing in enhancing the quality of generated responses. In Tables \ref{tab:mtcr}, a similar trend emerges. ThoT retains its lead, suggesting that its detailed prompt structure, which encourages summarizing and analyzing information in a structured manner, is particularly effective in complex conversational contexts. It underscores the importance of a methodical breakdown of context in generating relevant, accurate, and persona-consistent responses. The structured approach of ThoT prompts, which guide the model through a detailed, step-by-step analysis, consistently yields the best performance across chaotic contexts. 

\begin{table*}[!t]\small
\centering
\begin{tabular}{clc}
\toprule
\textbf{No.}        & \multicolumn{1}{l}{\textbf{Template}}                                                                        & \textbf{EM}           \\ \midrule
1                   & Let's read   through the document section by section, analyzing each   part   carefully as we go.             & 0.43                  \\ 
2                   & Take me   through this long document step-by-step, making sure not to   miss   any important details.         & 0.47                  \\ 
\multirow{2}{*}{3}  & Divide the   document into manageable parts and guide me through each   one,   providing insights as we move  & \multirow{2}{*}{0.51} \\ 
                    & along.                                                                                                        &                       \\ 
4                   & Analyze this   extensive document in sections, summarizing each one   and   noting any key points.            & 0.47                  \\ 
5                   & Let's go   through this document piece by piece, paying close attention   to   each section.                  & 0.50                  \\ 
6                   & Examine the   document in chunks, evaluating each part critically   before   moving to the next.              & 0.49                  \\ 
7                   & Walk me   through this lengthy document segment by segment, focusing   on   each part's significance.         & 0.52                  \\ 
8                   & Let's dissect   this document bit by bit, making sure to understand   the   nuances of each section.          & 0.45                  \\ 
9                   & Systematically   work through this document, summarizing and analyzing   each   portion as we go.             & 0.45                  \\ 
\multirow{2}{*}{10} & Navigate through this long document by breaking it   into smaller parts and   summarizing each, so we don't   & \multirow{2}{*}{0.48} \\ 
                    & miss   anything.                                                                                              &                       \\ 
11                  & Let's explore   the context step-by-step, carefully examining each segment.                                   & 0.44                  \\ 
12                  & Take me   through the context bit by bit, making sure we capture   all   important aspects.                   & 0.49                  \\ 
13                  & Let's   navigate through the context section by section, identifying   key   elements in each part.           & 0.47                  \\ 
14                  & Systematically   go through the context, focusing on each part   individually.                                & 0.46                  \\ 
15                  & Let's dissect   the context into smaller pieces, reviewing each one for   its   importance and relevance.     & 0.47                  \\ 
16                  & Analyze the   context by breaking it down into sections, summarizing   each   as we move forward.             & 0.49                  \\ 
17                  & Guide me   through the context part by part, providing insights along   the   way.                            & 0.52                  \\ 
18                  & Examine each   segment of the context meticulously, and let's discuss   the   findings.                       & 0.44                  \\ 
19                  & Approach the   context incrementally, taking the time to understand   each   portion fully.                   & 0.42                  \\ 
20                  & Carefully   analyze the context piece by piece, highlighting   relevant   points for each question.           & 0.47                  \\ 
21                  & In a   step-by-step manner, go through the context, surfacing   important   information that could be useful. & 0.53                  \\ 
22                  & Methodically   examine the context, focusing on key segments that   may   answer the query.                   & 0.45                  \\ 
23                  & Progressively   sift through the context, ensuring we capture all   pertinent   details.                      & 0.46                  \\ 
24                  & Navigate   through the context incrementally, identifying and   summarizing   relevant portions.              & 0.48                  \\ 
25                  & Let's   scrutinize the context in chunks, keeping an eye out   for   information that answers our queries.    & 0.42                  \\ 
26                  & Take a   modular approach to the context, summarizing each part   before   drawing any conclusions.           & 0.47                  \\ 
27                  & Read the   context in sections, concentrating on gathering insights   that   answer the question at hand.     & 0.48                  \\ 
\multirow{2}{*}{28} & Proceed through the context systematically,   zeroing in on areas that   could provide the answers we're      & \multirow{2}{*}{0.49} \\ 
                    & seeking.                                                                                                      &                       \\ 
\multirow{2}{*}{29} & Let's take a segmented approach to the context,   carefully evaluating each   part for its relevance to the   & \multirow{2}{*}{0.39} \\ 
                    & questions   posed.                                                                                            &                       \\ 
30                  & Walk me   through this context in manageable parts step by   step,   summarizing and analyzing as we go.      & \textbf{0.55}         \\ \bottomrule
\end{tabular}
\caption{Prompt Selection Analysis.}
\label{tab:middle_entityq}
\end{table*}

\subsection{Lost in Middle}
As shown in Table \ref{tab:middle}, we delves into the phenomena termed ``Lost in Middle'' \cite{Liu2307LostinMiddle}, where the focus is to examine the performance of various models on two different question-answering datasets, PopQA and EntityQ. The presented results draw a comparison between four methodologies: Vanilla, Retrieval, Chain of Thought (CoT), and Theory of Mind (ThoT), as applied to three advanced language models: GPT-4, GPT-3.5-turbo, and LLaMA 2 Chat (70B).

\paragraph{Performance on PopQA}: The results indicate that ThoT significantly outperforms the other methods across all three models. With GPT-4 leading at a score of 0.651, closely followed by GPT-3.5-turbo and LLaMA 2 Chat (70B) at 0.674 and 0.663, respectively. This suggests that ThoT's advanced technique, potentially incorporating more nuanced understandings of context and reasoning, has a definitive edge in handling the complexities of PopQA. The Vanilla approach yields moderate performance with GPT-4, which surpasses the scores of the other two models, hinting at the superior reasoning capabilities of the latest model iteration.

\paragraph{Performance on EntityQ}: Similar to PopQA, the ThoT methodology again tops the charts, indicating its robustness across different datasets. GPT-4's performance, while still the highest in the Vanilla method, sees a significant jump to 0.643 when applying ThoT, suggesting a better synergy between GPT-4's capabilities and ThoT's advanced reasoning framework. Notably, the Retrieval method showcases a stark improvement over Vanilla for all models, with LLaMA 2 Chat (70B) achieving the highest score of 0.643. 

\subsection{Impact of Model Scale}
\begin{figure}[t]
    \centering
    \begin{subfigure}{0.49\linewidth}
        \centering
        \includegraphics[width=\textwidth]{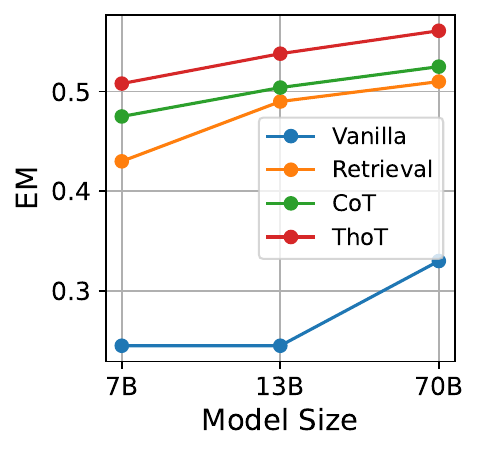}
        \caption{LLaMA 2}
        \label{fig:subfig1}
    \end{subfigure}
    \begin{subfigure}{0.49\linewidth}
        \centering
        \includegraphics[width=\textwidth]{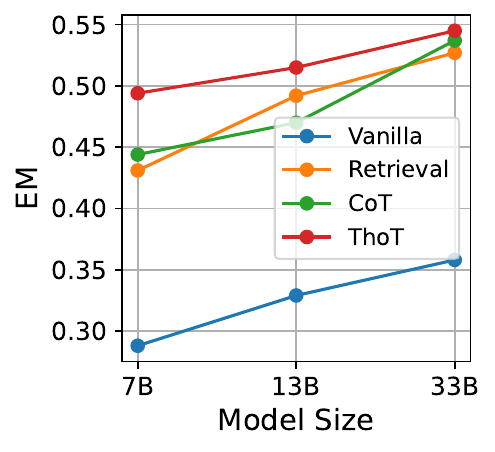}
        \caption{Vicuna}
        \label{fig:subfig2}
    \end{subfigure}
    \caption{PopQA performance on different scales of LLaMA 2 and Vicuna.}
    \label{fig:whole}
    \vspace{-5pt}
\end{figure}
As shown in Figure \ref{fig:whole}, results demonstrate a clear correlation between the scale of the model and its performance across different prompting strategies. As we scale up from 7 billion parameters to 70 billion parameters in the LLama2, there is a noticeable increase in the EM scores across all prompt configurations. The same trend is observed with the Vicuna model. This increase corroborates the hypothesis that larger models have a greater capacity for understanding and generating accurate responses. The Thought-by-Thought (ThoT) configuration maintains a lead in performance at all model sizes, which demonstrates ThoT prompts appear to leverage the model's capabilities more efficiently, regardless of model size. In addition, the experimental data suggests that model scale positively impacts the performance of retrieval-augmented generation methods, with prompts such as CoT and ThoT showing the most significant benefits. This trend underscores the importance of prompt design in harnessing the full potential of large language models. 

\subsection{Prompt Selection}
\label{subsec:prompt_selection}
\begin{figure*}[!t]
    \centering
    \includegraphics[width=1\linewidth]{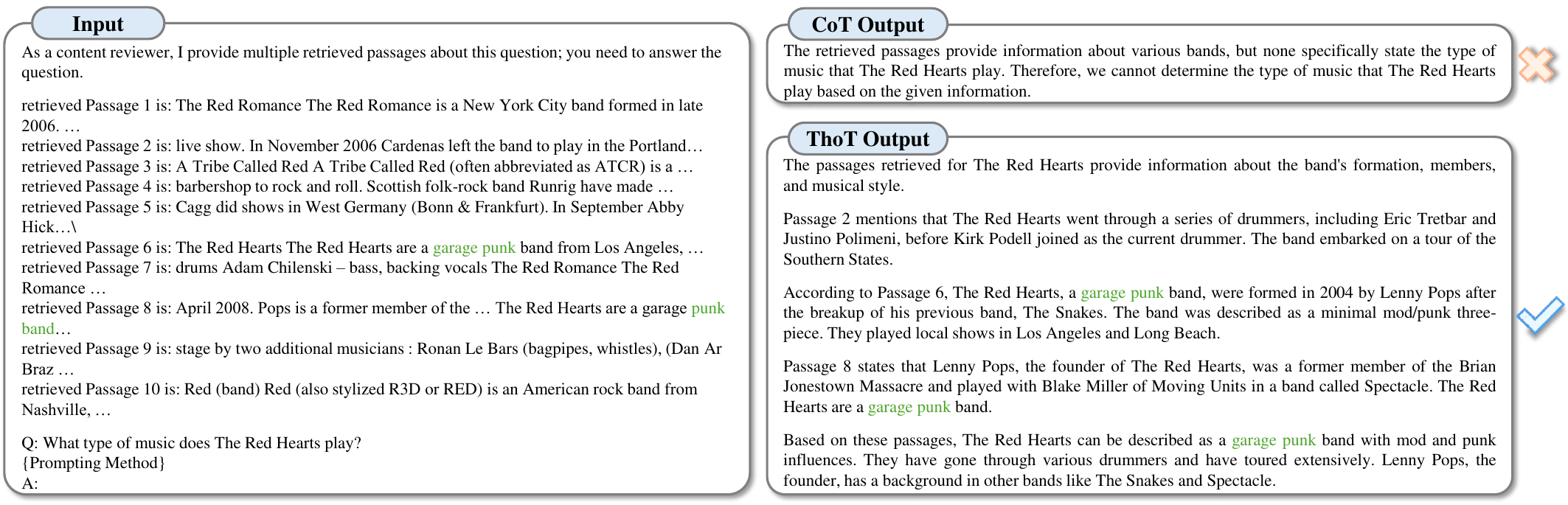}
    \caption{\small Case Study.}
    \label{fig:case}
\end{figure*}
\begin{figure*}[!t]
    \centering
    \includegraphics[width=1\linewidth]{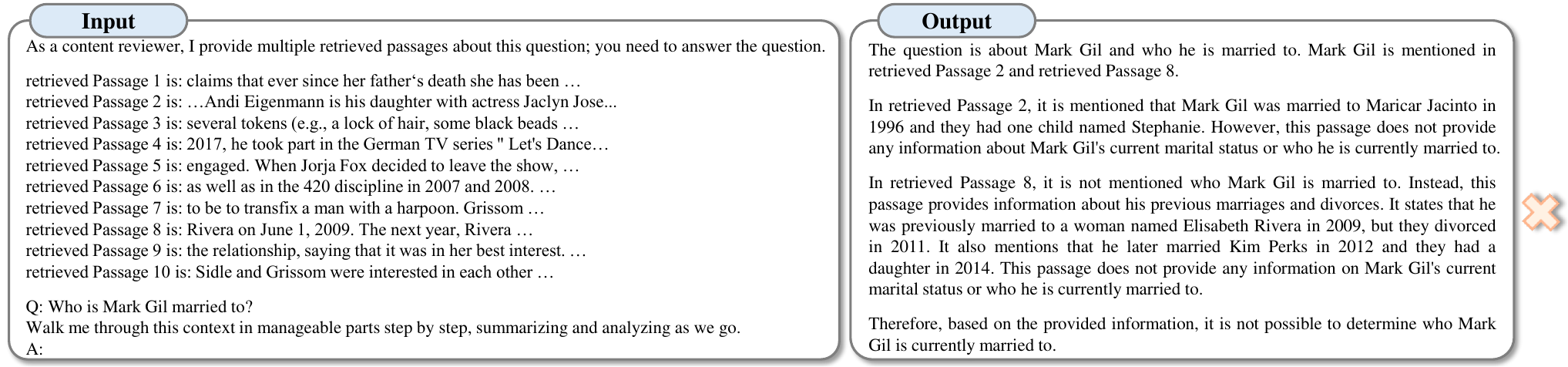}
    \caption{\small Error Analysis.}
    \label{fig:error}
\end{figure*}
The prompt analysis, as depicted in Table \ref{tab:middle_entityq}, reveals strategic variations in phrasing that have a direct impact on the performance of the language models, as indicated by the Exact Match (EM) scores. The prompts are designed to guide the model through the process of analyzing a document or context systematically and step by step.
A common theme among the more successful prompts, particularly those with EM scores above 0.50, is the explicit instruction to the model to not only analyze each part of the document but also to summarize and note key insights as it progresses. For instance, prompts that instructed the model to summarize each section and not miss important details, such as prompt 2 and prompt 4, resulted in higher EM scores.
Prompts that encouraged a more granular approach, directing the model to focus on individual parts and their significance or relevance, also performed well. This is evidenced by prompt 14, which achieved a relatively high EM score. The more detailed the instruction for the model to dissect and analyze the context, the better the model performed.
Conversely, prompts that were less directive or less structured, such as prompt 29, tended to result in lower EM scores. This suggests that models benefit from clear, specific, and action-oriented instructions that leave little room for ambiguity in the analytical process.
The highest-scoring prompt, number 30, combines several elements of successful prompts. It asks the model to manage the complexity by breaking it down into parts, which implies a thorough analysis, and also to summarize and analyze, indicating an active engagement with the material that goes beyond mere reading or passive understanding.
In summary, the results suggest that prompts that are structured to enforce a detailed analytical process, encouraging step-by-step dissection, summarization, and critical evaluation, lead to better model performance. 

\subsection{Case Study}
The case study presented in Figure \ref{fig:case} shows a comparative analysis between the CoT and ThoT in PopQA. CoT only stated that the passages contained information about various bands without specifying the genre of ``The Red Hearts''. This illustrates a potential limitation of the CoT approach: it might not effectively synthesize information from multiple sources when the answer is not explicitly stated but rather needs to be inferred from the given data. On the contrary, the ThoT method successfully identified that ``The Red Hearts play garage punk music''. This outcome showcases the strength of the ThoT approach. ThoT is adept at synthesizing and correlating information across multiple pieces of text. It pieced together relevant details from passages 6 and 8, noting that ``The Red Hearts'' were described as ``a garage punk band''.

\subsection{Error Analysis}
From Figure \ref{fig:error}, the ThoT method can not conclude the answer for this case. The passage stating, ``Andi Eigenmann is his daughter with actress Jaclyn Jose'' holds the key to the correct inference that Mark Gil was married to Jaclyn Jose. The ThoT method's failure to make this inference suggests that while the model is adept at extracting explicit information, it struggles with implicit reasoning that requires understanding nuanced relationships. The oversight may be attributed to the model's inferential reasoning capabilities, specifically regarding relationship inference—a known shortcoming in large models as also identified in prior research \cite{Berglund2309Reversal}. The case study highlights the need for models to not only parse and summarize information but also engage in a level of deductive reasoning that resembles human cognition. Therefore, enhancing the model's ability to infer and reason about entity relationships is very important.

\section{Conclusion}
This paper presented the ``Thread of Thought'' (ThoT) strategy, a novel approach designed to enhance the performance of Large Language Models (LLMs) in processing chaotic contextual information. ThoT, inspired by human cognitive processes, significantly improves the ability of LLMs to segment and analyze extended contexts. We compared ThoT with existing methods, which often require complex retraining, fine-tuning, or are limited in their ability to handle large volumes of intricate information. ThoT, in contrast, offers a more straightforward and efficient solution. It acts as a ``plug-and-play'' module, seamlessly integrating with various pre-trained language models and prompting strategies without necessitating complex procedures. The effectiveness of ThoT was rigorously tested using long-tail question answering datasets, such as PopQA and EntityQ, and a Multi-Turn Conversation Response dataset based on everyday conversations. The results from these evaluations were clear: ThoT not only excelled in handling chaotic contexts but also enhanced the reasoning capabilities of LLMs.

\bibliography{anthology,custom}
\end{document}